\documentclass[review]{elsarticle}
\usepackage[utf8]{inputenc}
\usepackage{lineno,hyperref}
\usepackage{tikz}
\usetikzlibrary{shapes.geometric, arrows}
\usepackage{algorithm}
\usepackage{algorithmic}
\usepackage[top=1.65in, bottom=1.65in, left=1.35in, right=1.35in]{geometry}
\usepackage{setspace}
\usepackage{amsmath}
\usepackage{tcolorbox}
\usepackage{subcaption}
\usepackage{upquote}

\usepackage{pgfplots}
\pgfplotsset{width=7cm,compat=1.15}
\usepgfplotslibrary{statistics}

\usepackage{caption} 
\newtcolorbox[auto counter]{example}[1][]{
  fonttitle=\scshape,
  title={Example \thetcbcounter},
  #1
}

\journal{Applied Soft Computing -- arXiv.2307.15464}
\biboptions{numbers}

\makeatletter
\def\ps@pprintTitle{%
 \let\@oddhead\@empty
 \let\@evenhead\@empty
 \def\@oddfoot{}%
 \let\@evenfoot\@oddfoot}
\makeatother
\begin{document}

\begin{frontmatter}

\title{Automatic Design of Semantic Similarity Ensembles Using Grammatical Evolution}

\author{Jorge Martinez-Gil}
\address{Software Competence Center Hagenberg GmbH \\ Softwarepark 32a, 4232 Hagenberg, Austria \\ \url{jorge.martinez-gil@scch.at}}

\begin{abstract}
Semantic similarity measures are a key component in natural language processing tasks such as document analysis, requirement matching, and user input interpretation. However, the performance of individual measures varies considerably across datasets. To address this, ensemble approaches that combine multiple measures are often employed. This paper presents an automated strategy based on grammatical evolution for constructing semantic similarity ensembles. The method evolves aggregation functions that maximize correlation with human-labeled similarity scores. Experiments on standard benchmark datasets demonstrate that the proposed approach outperforms existing ensemble techniques in terms of accuracy. The results confirm the effectiveness of grammatical evolution in designing adaptive and accurate similarity models.
\end{abstract}


\end{frontmatter}

\section{Introduction}
In recent times, ensemble learning has become a widely used technique to address the limitations of individual methods by aggregating them into a unified model. Using the predictions of diverse methods aims to mitigate individual method shortcomings, such as outliers in response to specific inputs. Therefore, the fundamental premise behind ensemble learning is the expectation that a carefully chosen set of methods will yield superior results compared to any single method alone \cite{key-superlearning}.

While ensemble learning has attracted considerable attention and received extensive research efforts \cite{key-AutoML}, its application in semantic similarity measurement remains largely unexplored. This presents an opportunity to show the potential of this approach to address the challenge of automatically determining semantic similarity between pieces of textual information. The reason is that, despite advancements in semantic similarity measures, a lack of consensus persists among the individual suitability of these measures when assessing the semantic similarity between textual information \cite{key-Harispe}.

Programming languages have structured syntax and semantics that can be used to build ensembles. Grammatical evolution takes advantage of the formal grammar of programming languages to automate the design of semantic similarity measure ensembles. The motivation behind this approach comes from the idea that a diversified pool of semantic similarity measures can compensate for the inherent limitations of individual measures \cite{key-Ballatore}. Through the aggregation of multiple measures, our proposed approach seeks to benefit from the diversity of these measures to achieve a higher level of agreement.

Through this research, we aim to contribute to natural language processing (NLP) by providing a novel perspective on semantic similarity measurement. We propose adopting Grammatical Evolution (GE) \cite{key-grammatical0} as an ensemble learning strategy to address the misalignment among existing semantic similarity measures. Empirical evaluations conducted on three well-known benchmark datasets will demonstrate the effectiveness of GE ensemble-based approaches in improving performance concerning most existing methods' capabilities.

The rationale behind this research is that GE can bring a new point of view to the semantic similarity measurement domain. The collective recommendation capability of various similarity measures allows for augmenting the quality of semantic similarity assessments, paving the way for more reliable real-world applications. Therefore, the major contributions of this work can be summarized as follows:

\begin{itemize}
	\item We propose, for the first time, the automatic learning of semantic similarity ensembles based on the notion of GE. This method offers advantages such as high accuracy, excellent interpretability, a platform-independent solution, and easy transferability to problems of analog nature.
	\item We implement and empirically evaluate our strategy to compare it with existing work and demonstrate its superiority in solving some of the most well-known dataset benchmarks used by the research community.
\end{itemize}

The rest of this paper is organized as follows: Section 2 provides an overview of related work in ensemble learning using GE and other kinds of ensembles for semantic similarity. Section 3 introduces the problem statement. Section 4 presents the details of the proposed GE strategy to address the challenge. Section 5 describes the experimental setup and presents the evaluation results. Section 6 discusses the results obtained and future work directions. Finally, Section 7 concludes the paper.

\section{State-of-the-art}
GE is a particular form of genetic programming (GP) that uses a formal grammar (FG) to generate computer programs \cite{key-grammatical0}. GE is considered an evolutionary strategy that makes use of a genotype-to-phenotype strategy. To do that, GE uses an FG definition to describe the language that the model might produce. The most common approach uses the Backus-Naur Form (BNF) \cite{key-bnf}, a widely used notation to formulate an FG using production rules. These rules include terminals and non-terminals (which can be expanded into terminal and non-terminal symbols).

The BNF grammar allows defining the structure of the ensembles to be learned. Please note that in this work, the term ensemble is equivalent to a program aiming to aggregate an initial set of semantic similarity measures as effectively and efficiently as possible. The FG acts, therefore, as the guideline for the evolution of the ensembles, and it defines the set of valid ensembles that can be generated. This allows for a more controlled evolution compared to rival techniques.

Furthermore, the evolution of the learning process is guided towards optimizing a fitness function, which measures the quality of the generated ensembles in the training phase. In our case, we can evaluate the quality based on the degree of correlation it presents concerning human judgment. Moreover, this fitness function also allows the selection of the ensembles that will be used as the parents in the next generation. This process is repeated until a good enough solution has been reached or a pre-defined number of iterations has been consumed.

Apart from the possibility of reaching high degrees of accuracy, the other significant advantage of this approach is the ability to generate models that adhere to a specific syntax and structure (i.e., good interpretability of the resulting models). Therefore, this approach is advantageous in domains where the capability of understanding the solution is essential.

\subsection{Semantic Similarity}
The challenge of semantic similarity measurement is a critical task in many computer-related fields \cite{key-sts,key-Lastra2,key-martinez-eswa,key-Pedersen,key-Pirro2,key-Rus}. It aims to quantitatively capture the degree of likeness between two pieces of text based on their underlying meaning \cite{key-Lastra-HESML}. In recent years, significant progress has been made in this field, leading to the development of state-of-the-art techniques \cite{key-martinez-eswa2}. One prominent approach involves utilizing deep learning (DL) models, such as transformer-based architectures like BERT (Bidirectional Encoder Representations from Transformers) \cite{key-Bert}. These models are pre-trained on vast amounts of text, enabling them to learn text representations \cite{key-martinez-datak}. Fine-tuning these models has shown remarkable performance, outperforming traditional methods that rely on handcrafted features \cite{key-Navigli}.

Another line of research focuses on using distributional semantics, which captures meaning using distributional patterns of words in a large corpus. Methods such as word embeddings (e.g., Word2Vec \cite{key-Mikolov}) represent words as vectors in a continuous vector space. The semantic resemblance between the textual pieces can then be estimated by comparing the vector representations of these pieces using methods like cosine distance. Additionally, recent studies have explored incorporating contextual information using contextualized word embeddings, such as Embeddings from Language Models (ELMo) \cite{key-elmo} and Universal Sentence Encoder (USE) \cite{key-cer}. Considering the surrounding words, these models generate context-dependent word representations, leading to improved semantic similarity estimation in a given context.

In recent times, ensembles have also emerged as a helpful technique in semantic similarity measurement, offering a reasonable solution to the challenges posed by the inherent complexity of human language \cite{key-martinez-ijseke}. The idea of aggregating multiple semantic similarity measures allows ensembles to mitigate the limitations of individual measures and capture a better understanding of semantic similarity \cite{key-Bar}. Ensembles exploit each measure's inherent complementarity and different perspectives by using the diversity of these existing measures \cite{key-Potash}. Improving performance and transfer learning capabilities is usually possible \cite{key-martinez-transfer}. With their ability to aggregate diverse perspectives and mitigate model biases, ensembles have proven helpful in semantic similarity measurement, pushing the boundaries of accuracy and offering promising lines of research \cite{key-martinez-mlwa}.

In summary, state-of-the-art techniques for semantic similarity measurement have witnessed significant progress in the last years, driven by the use of DL models, the incorporation of contextual information, and the exploitation of ensembles. These approaches have demonstrated exemplary performance, being superior to traditional methods. As the field continues to evolve, further research and development are expected to improve the existing methods, facilitating many computer-related applications.

\subsection{Grammatical evolution}
GE is a well-known technique in the domain of GP, combining the principles of genetic algorithms (GAs) and FGs. It has gained recognition as a state-of-the-art approach for evolving computer programs that exhibit complex behaviors \cite{key-grammatical}. It offers a framework to automatically generate programs (ensembles in our particular case) by evolving their syntax and semantics through a GA. 

The ensembles can be represented through strings of symbols, which allows their manipulation and evolution using genetic operators through FGs. This facilitates the exploitation of a vast search space that allows the discovery of practical solutions to a wide range of computational problems \cite{key-whigham}. Over time, GE has undergone remarkable advances, including knowledge integration, mutation process improvements, and new crossover operators. These advances have improved the accuracy and scalability of GE-based solutions, making it one of the most promising techniques in the GP landscape \cite{key-wang}.

The state-of-the-art in GE involves developing hybrid approaches that combine GE with other techniques like particle swarm optimization \cite{key-kennedy}. These hybrid approaches use the strengths of multiple techniques to overcome limitations and improve search capability. Additionally, increased focus is on improving scalability through parallel and distributed computing paradigms. Researchers have been able to solve some computationally intensive problems using these paradigms. Furthermore, advancements in fitness approximation techniques have significantly improved efficiency by reducing computational overhead. Continually exploring novel techniques aims to improve this GP approach's performance.

\subsection{Differences between Genetic Programming and Genetic Algorithm}
The main distinction between GP and GAs is their optimization approaches. GAs optimize a given function by searching for optimal parameter values, while GP generates programs (ensembles in this case) that perform well on a specific task. GP uses a higher-level representation to capture complex relationships among variables, enabling the encoding of complex solutions within the population. It incorporates a refined selection process to maintain population diversity and avoid premature convergence. GP's crossover operator generates novel solutions, while its mutation operator maintains diversity by introducing variations. Additionally, GP utilizes a complex fitness function that ensures a thorough assessment of ensembles during the evolutionary process.

\subsection{Contribution over the state-of-the-art}
We propose exploring GE as a suitable approach for learning ensembles within the domain of semantic similarity measurement. The main goal is to identify a program that achieves a near-optimal fitness value for a given objective function to emulate human judgment. While traditional methods often rely on tree-structured expressions for direct manipulation \cite{key-martinez-ijseke}, our approach applies genetic operators to an integer string, which is then converted into an ensemble using a BNF grammar. Although this paper does not focus on this aspect, the same strategy could be extended to identify source code clones \citep{key-martinez-clones}. 

This approach offers several benefits, including higher accuracy, improved interpretability of the resulting models, and easier translation of the models into widely used programming languages. Moreover, unlike traditional ensemble methods such as boosting or bagging, GE allows for the dynamic evolution of ensemble structures without predefined aggregation rules, and unlike neural ensembles, GE does not require extensive training data or GPU-based computation, making it more suitable for low-resource settings.

\section{Problem Statement}
Let us assume that we have a set of candidate similarity measures $\mathcal{M} = {M_1, M_2, ..., M_n}$, where $n$ is the total number of candidates. Let us assume that each $M_i$ takes a pair of textual pieces $X$ and $Y$ as input and produces a similarity score $S_i$ as output.

We aim to automatically select a subset from $\mathcal{M}$ and aggregate them into an ensemble $E$, such that $E(X,Y)$ provides an accurate semantic similarity score.

Let us also assume that we have a vector $\mathbf{w} = [w_1, w_2, ..., w_n]$, where $w_i \in {0, 1}$ represents the inclusion of $M_i$ in the ensemble. If $w_i = 1$, then $M_i$ is selected; otherwise, if $w_i = 0$, $M_i$ is excluded from $E$.

The ensemble function $E(X,Y)$ is defined as the aggregation of a subset from $\mathcal{M}$ where the measures are weighted by their corresponding aforementioned inclusion values as shown in Eq. \ref{eq:e}:

\begin{equation}
	E(X, Y) = \sum_{i=1}^{n} w_i \cdot M_i(X, Y)
	\label{eq:e}
\end{equation}

In this research, we use GE to build the ensemble function. Please note that GE provides a framework for generating and evolving an ensemble based on BNF grammar. In this case, the BNF grammar defines the rules for building the aggregation strategies.

Therefore, the problem consists of finding the $\mathbf{w}$ that maximizes the ensemble's performance. Examples of performance can be measures such as precision and recall. Nevertheless, in the case of semantic similarity measurement, the challenge is to emulate human judgment \cite{key-Ballatore}. This means that we need to use methods such as correlation coefficients. Therefore, we aim to optimize the correlation between the ensemble results and a human-curated ground truth dataset.

To do that, given a gold standard $\mathcal{G}$, i.e., a dataset created and curated by human experts, the goal is to maximize the correlation between the $\mathcal{G}$ and the results from the proposed strategy $\mathcal{S}$ as shown in Eq. \ref{eq:max}.

\begin{equation}
S={arg \ max}_{S} \ correl(\vec{\mathcal{G}},\vec{\mathcal{S}})
\label{eq:max}
\end{equation}

$\mathcal{S}$ can take different semantic similarity measures as input. These measures will function as weak estimators to obtain intermediate semantic similarity scores to learn a higher-level yet robust strategy able to work over unseen data. In short, the goal is to identify an ensemble capable of adapting to training data and performing well on data never seen before.

GE can evolve candidate ensembles, evaluating their correlation to a human-curated training set. The fitness function guides the search process by assigning fitness to each candidate ensemble based on performance. The process iteratively evolves the population of candidate ensembles, using genetic operators, until a termination condition is met, such as reaching a maximum number of generations (previously defined by the operator) or achieving a satisfactory fitness level for the problem at hand, since the ideal result will be challenging to achieve.

In this way, a computer language's syntax and semantics can be created following the rules described within GE. These criteria are applied to produce a population of computer programs, or ensembles in our specific case, capable of evolving. The approach generates new strings of symbols equivalent to the most successful ensembles in the population. 

The degree to which an ensemble successfully correlates to the ground truth is critical in determining that success. The reason is that ensembles that are more successful at completing a test have a greater chance of being picked for reproduction and mutation. In contrast, the less successful ones will not be passed on to the next generation. The rationale behind this approach is that the population changes over time, with more successful ensembles becoming prevalent. 

One of the most significant benefits is that GE facilitates building ensembles that can solve complex issues automatically. During the evolutionary process, the approach automatically explores the space of possible ensembles and selects the one that maximizes the performance. Our hypothesis is that the resulting ensemble can estimate semantic similarity for unseen textual inputs. This hypothesis will be empirically tested later in this paper.

\section{Methods}
We have seen that GE is a powerful evolutionary computation technique that combines GAs with an FG. We can automatically learn complex similarity models capable of capturing the nuances of natural language by using the adaption capability of GE.

The process begins with the definition of a BNF grammar that represents the structure of the possible semantic similarity models. This BNF grammar serves as a guideline for generating diverse candidate solutions. Each candidate solution represents a unique ensemble of semantic similarity measures. Algorithm \ref{algorithm:grammatical-evolution} shows us how, through an iterative process, the approach explores the space of potential solutions, gradually improving their performance through fitness evaluation and selection.

\begin{algorithm}
\caption{Grammatical Evolution using Genetic Programming}
\label{algorithm:grammatical-evolution}
\begin{algorithmic}[1]
  \STATE \textbf{Input:} Grammar $G$, Population size $N$, Termination condition
  \STATE \textbf{Output:} Best individual
  
  \STATE Initialize population $P$ with $N$ random individuals
  \STATE Evaluate fitness for each individual in $P$

  \WHILE{termination condition not met}
    \STATE Select parents for reproduction based on fitness
    \STATE Initialize empty offspring population $O$

    \FOR{each pair of parents}
      \STATE Apply crossover to create two offspring
      \STATE Apply mutation to each offspring
      \STATE Add the offspring to $O$
    \ENDFOR

    \STATE Evaluate fitness for the offspring in $O$
    \STATE Select individuals for the new population based on fitness
    \STATE Replace the current population $P$ with the new population
  \ENDWHILE

  \RETURN Best individual
\end{algorithmic}
\end{algorithm}

The fitness evaluation is based on an objective function that measures the quality of the ensembles. This function could consider factors such as the ensemble's output's accuracy, or diversity, although this research focuses on accuracy. The idea behind aggregating multiple semantic similarity measures allows the ensembles to capture different aspects of the problem. The adaptive nature of the process enables the ensembles to learn and evolve, continuously refining their performance over time. 

GE not only automates the ensemble learning process but also pushes the boundaries of semantic similarity modeling. Allowing the ensembles to learn from data eliminates the need for manual feature engineering (e.g., manual selection of similarity measures), which can be time-consuming and error-prone. Instead, the ensembles adapt to the training data, uncovering hidden patterns that may not be apparent to the human eye.

\subsection{Mathematical Foundation}
GE performs search over a space of programs via a genotype-to-phenotype mapping guided by a formal grammar.

\paragraph{Genotype}
Let the genotype be a binary string $G \in \{0,1\}^n$, partitioned into $k$ codons (substrings), i.e.,
\[
G = (g_1, g_2, \dots, g_k), \quad g_i \in \{0,1\}^{\ell}, \; \ell \text{ fixed}
\]
Each codon $g_i$ is interpreted as an integer $c_i \in \mathcal{N}$ via binary-to-decimal conversion.

\paragraph{Grammar}
Let $\mathcal{C} = (N, T, R, S)$ be a context-free grammar in Backus-Naur Form (BNF), where:
\begin{itemize}
  \item $N$ is the set of non-terminals,
  \item $T$ is the set of terminals,
  \item $R$ is the set of production rules $A \rightarrow \alpha$ with $A \in N$, $\alpha \in (N \cup T)^*$,
  \item $S \in N$ is the start symbol.
\end{itemize}

\paragraph{Mapping Function}
The mapping function $f: \{0,1\}^n \to (N \cup T)^*$ produces a phenotype $P = f(G)$ by recursively applying production rules from $\mathcal{C}$ using codons $c_i$:
\[
f(G) = \text{derivation}(S, C), \quad C = (c_1, \dots, c_k)
\]
At each step, the next codon $c_i$ selects among the $r$ available expansions for the current non-terminal $A$:
\[
A \rightarrow \alpha_{(c_i \bmod r)}
\]
The process continues until all non-terminals are expanded or the codon list is exhausted.

\paragraph{Output}
The final phenotype $P$ is a syntactically valid program (expression tree or string) derived entirely in terminal symbols, i.e., $P \in T^*$.

\paragraph{Summary}
GE thus defines a deterministic but grammar-constrained mapping from binary strings to executable programs:
\[
f: \{0,1\}^n \rightarrow T^*
\]
guided by codon-driven rule selection within a formal grammar $\mathcal{C}$.

\subsection{Fitness Function}
Let $F(w)$ represent the fitness function that evaluates how well an ensemble, defined by the vector $w$, performs on a semantic similarity task. This function is based on a performance metric, such as a correlation coefficient:

\[F(w) = \rho(y, \hat{y}(w)) \]

where:
\begin{itemize}
    \item $y$ is the set of ground-truth similarity scores.
    \item $\hat{y}(w)$ refers to the predicted similarity scores from the ensemble defined by $w$.
    \item $\rho$ is a correlation coefficient.
\end{itemize}

\subsection{Genetic Operators}
Genetic operators modify ensemble configurations during the evolution process. Two primary operators are crossover and mutation.

\subsubsection{Crossover} 
Crossover combines two parent ensembles, $w^1$ and $w^2$, to produce offspring. In a one-point crossover mechanism:
\[
w^{1'} = [w^1_1, w^1_2, \ldots, w^1_k, w^2_{k+1}, \ldots, w^2_d]
\]
\[
w^{2'} = [w^2_1, w^2_2, \ldots, w^2_k, w^1_{k+1}, \ldots, w^1_d]
\]
where $k$ is the crossover point, and $d$ is the length of the vector.

\subsubsection{Mutation} 
Mutation introduces variation by modifying elements of $w$. Specifically, positions in $w$ are randomly selected, and their values are flipped. Mathematically, this is expressed as:
\[
w_i' =
\begin{cases} 
w_i, & if \ r > p_mut, \\
1 - w_i, & if \ r \leq p_mut,
\end{cases}
\]
where:
\begin{itemize}
    \item $w_i$ is the $i$-th element of $w$.
    \item $p_mut$ is the mutation probability.
    \item $r$ is a random number uniformly sampled from $[0, 1]$.
\end{itemize}

\subsection{Grammar Rules}
Phenotype generation is constrained by a context-free grammar $\mathcal{C} = (N, T, R, S)$, where:

\begin{itemize}
  \item $N$ is the set of non-terminal symbols,
  \item $T$ is the set of terminal symbols,
  \item $R$ is the set of production rules $A \rightarrow \alpha$, with $A \in N$, $\alpha \in (N \cup T)^*$,
  \item $S \in N$ is the start symbol.
\end{itemize}

At each derivation step, codon $c_i$ determines which rule to apply for a non-terminal $A$ among its $r$ possible expansions:
\[
A \rightarrow \alpha_{(c_i \bmod r)}
\]
This ensures syntactic correctness of the resulting phenotype and allows enforcement of structural constraints through $\mathcal{C}$.

\paragraph{Domain-Specific BNF}
To tailor the search space to semantic similarity tasks, a grammar $\mathcal{C}$ must encode the domain-specific operations permissible in ensembles. This typically includes numeric aggregation, transformation functions, and feature access. A simplified Python-inspired fragment is:

\begin{example}[label=ex:01]
\begin{tcolorbox}
\begin{verbatim}
<expr>  ::=  <expr>+<expr> |
             <expr>-<expr> |
             <expr>*<expr> |
             pdiv(<expr>,<expr>) |
             psqrt(<expr>) |
             np.sin(<expr>) |
             np.tanh(<expr>) |
             np.exp(<expr>) |
             plog(<expr>) |
             x[:,0] | x[:,1] | ... | x[:,4] |
             <c><c>.<c><c>

<c>     ::= 0 | 1 | 2 | 3 | 4 | 5 | 6 | 7 | 8 | 9
\end{verbatim}
\end{tcolorbox}
\end{example}

\paragraph{Representation and Operator Support}
Unlike traditional Genetic Programming (GP), GE permits manipulation at multiple levels:
\begin{itemize}
  \item Genotype level: binary or integer codons,
  \item Partial phenotypes: derivation trees in progress,
  \item Full phenotypes: completed executable programs.
\end{itemize}
This flexibility broadens the search space while maintaining syntactic validity, which contributes to effective exploration and better convergence.

\paragraph{Implementation}
All experiments were implemented using the PonyGE2 framework~\cite{key-ponyge2}, which supports codon-based GE and customizable grammars. It facilitates reproducible evolutionary runs and integrates common genetic operators, grammar parsing, and evaluation infrastructure.

\section{Results}
In this section, we present the findings of our experiments focused on semantic similarity measurement. We will also explore two ways to build ensembles using the Python language. From now on, we will call one GE, which will only search for accuracy. Furthermore, the other, which we will call GE-i from now on, will look for a Python style that facilitates interpretability. We will see examples later and conduct a comparative analysis of the outcomes produced by our proposed strategies concerning state-of-the-art GP techniques.

\subsection{Empirical Setup and Baseline Selection} 
Table \ref{tab:param} presents our setup concerning the set of parameters and their corresponding values associated with the PonyGE2 framework \cite{key-ponyge2}. The technical details of each of the entries in the table are beyond the scope of this paper but can be consulted at \cite{key-software}. The purpose of this table is to provide a concise overview of the configuration settings used in the context of a particular study or experiment.

\begin{table}
  \centering
	\caption{Parameters that have been established for ensemble learning using GE}
\begin{tabular}{|l|l|}
\hline
\textbf{Parameter} & \textbf{Value} \\
\hline
CROSSOVER & variable\_onepoint \\
CROSSOVER\_PROBABILITY & 0.8 \\
GENERATIONS & 200 \\
MAX\_GENOME\_LENGTH & 1000 \\
INITIALISATION & PI\_grow \\
INVALID\_SELECTION & False \\
MAX\_INIT\_TREE\_DEPTH & 10 \\
MAX\_TREE\_DEPTH & 18 \\
MUTATION & int\_flip\_per\_codon \\
POPULATION\_SIZE & 100 \\
FITNESS\_FUNCTION & max \\
REPLACEMENT & generational \\
SELECTION & tournament \\
\hline
\end{tabular}
  \label{tab:param}
\end{table}

Please note that one-point crossover is a widely used method in evolutionary algorithms. It helps maintain diversity by exchanging genetic material between individuals while avoiding excessive disruption of well-performing solutions.

Our baseline is one of the top-performing methods for aggregating similarity scores, i.e., linear regression \cite{key-Lastra}. Linear regression aims to establish a functional relationship between the previously considered semantic similarity measures. This relationship can be represented using a mathematical equation, which connects the output with multiple semantic similarity measures, as depicted in Eq. \ref{eq:lr}.

\begin{equation}
    \vec{\hat{\alpha}} = arg \ min \left(D, \vec{\alpha}\right) = arg \ min \sum_{i=1}^{n} \left(\vec{\alpha} \cdot \vec{a_i} - b_i\right)^2
\label{eq:lr}
\end{equation}

Eq. \ref{eq:lr} represents the minimization problem involved in linear regression, aiming to find the optimal vector $\vec{\hat{\alpha}}$ that minimizes the discrepancy $D$ between the predicted values and the actual values. The optimization process seeks to minimize the sum of squared differences between the dot product of the vector $\vec{\alpha}$ and the vector $\vec{a_i}$, representing the semantic similarity measures, and the corresponding target values $b_i$. The symbol $arg \ min$ denotes the argument that minimizes the expression within the parentheses, and the index $i$ ranges from 1 to $n$, representing the number of instances. In this way, linear regression is a foundational approach for building ensembles by quantifying the association between the semantic similarity measures and the desired output, allowing for the derivation of predictive models.

\subsection{Datasets}
The first dataset used in our experiments is the so-called \textbf{Miller \& Charles} dataset \cite{key-Miller}, from now \textbf{MC30}. This is the standard dataset community members use when evaluating research methodologies that concentrate on general cases. It includes 30 use cases comparing words of daily use. Therefore, this dataset aims to evaluate the semantic similarity between words that are components of a general-purpose scenario.

The second dataset is the so-called \textbf{GeReSiD50} dataset \cite{key-Ballatore2} and is drawn from the realm of geospatial research. It covers a pool of textual phrases, each of which has been grouped into one of 50 unique pairings. This pool of sentences includes over 100 different geographical expressions. On each of the 50 pairings, human opinions about the degree of semantic similarity were solicited and recorded individually. These 50 pairings include samples that are in no way comparable to one another and others that, in human view, are virtually indistinguishable.

The third dataset is the so-called \textbf{WS353} dataset \cite{key-agirre}, a widely used benchmark for evaluating semantic similarity in NLP tasks. It consists of 353-word pairs, each annotated with human-assigned similarity scores, providing a reference for comparing computational models' performance in capturing word-level meaning.

\subsection{Evaluation Criteria}
Our goal is to measure the correlation of our results to human judgment. This is the standard procedure for measuring how accurately predicted semantic similarity aligns with reference values \cite{key-Lastra-HESML}. The Pearson Correlation Coefficient (PCC) and the Spearman Rank Correlation Coefficient (SRCC) are two commonly used metrics. The PCC evaluates the degree of linear association between predicted and reference values, focusing on proportional alignment. The SRCC, in contrast, assesses how well the predicted and reference rankings match, making it suitable for tasks where the order of similarity is more important than precise values. Together, these measures provide valuable quantitative feedback for assessing the performance of semantic similarity models. This study aims to closely examine the ensemble's accuracy concerning these two correlation coefficients, as discussed in \cite{key-Hill}. Please also note that even with small search spaces, GE is still more efficient than an exhaustive search for small search spaces because it explores solutions using evolutionary principles, reducing computational effort and time by avoiding a complete enumeration of possibilities.

\subsection{Empirical Results}
We provide an overview of the outcomes derived from our empirical assessment of the above benchmarks. Tables \ref{tab:data} and \ref{tab:data2} show the reference data for the semantic similarity measures that will be part of the ensemble for solving the \textbf{MC30} and the \textbf{GeReSiD50} benchmark datasets, respectively. Our primary pool of measures will be based on different variants over BERT \cite{key-Bert} since there is a broad consensus about their superiority in tackling this task. \textbf{Truth} represents the ground truth values, ranging from 0 to 1, as a reference for comparison. \textbf{Bert-Cos.} displays the results obtained by encoding the text pieces using BERT and calculating similarity based on the cosine formula. \textbf{Bert-Man.} presents results obtained using the Manhattan distance. \textbf{Bert-Euc.} shows results based on the Euclidean distance. \textbf{Bert-Inn.} reflects results obtained using the Inner Product similarity measure. Lastly, \textbf{Bert-Ang.} illustrates results obtained by calculating similarity using the cosine of the angle. 

\begin{table}[htbp]
  \centering
  \caption{Results obtained for the \textbf{MC30} benchmark dataset by different methods in isolation}
    \begin{tabular}{|c|c|c|c|c|c|c|}
    \hline
    \textbf{UC} & \textbf{Truth} & \textbf{Bert-Cos.} & \textbf{Bert-Man.} & \textbf{Bert-Euc.} & \textbf{Bert-Inn.} & \textbf{Bert-Ang.} \\
    \hline
    UC1 & 1.000 & 0.921 & 0.642 & 0.642 & 0.993 & 0.873 \\
    UC2 & 0.980 & 0.818 & 0.462 & 0.462 & 0.863 & 0.805 \\
    UC3 & 0.980 & 0.899 & 0.607 & 0.605 & 0.922 & 0.856 \\
    UC4 & 0.959 & 0.936 & 0.678 & 0.680 & 1.000 & 0.886 \\
    UC5 & 0.944 & 0.860 & 0.525 & 0.526 & 0.916 & 0.830 \\
    UC6 & 0.921 & 0.558 & 0.165 & 0.170 & 0.577 & 0.688 \\
    UC7 & 0.893 & 0.839 & 0.488 & 0.491 & 0.893 & 0.817 \\
    UC8 & 0.872 & 0.855 & 0.507 & 0.512 & 0.926 & 0.826 \\
    UC9 & 0.793 & 0.824 & 0.471 & 0.465 & 0.886 & 0.808 \\
    UC10 & 0.786 & 0.615 & 0.216 & 0.208 & 0.665 & 0.711 \\
    UC11 & 0.778 & 0.512 & 0.124 & 0.125 & 0.533 & 0.671 \\
    UC12 & 0.758 & 0.679 & 0.296 & 0.291 & 0.704 & 0.738 \\
    UC13 & 0.753 & 0.842 & 0.494 & 0.492 & 0.907 & 0.818 \\
    UC14 & 0.719 & 0.621 & 0.230 & 0.222 & 0.658 & 0.713 \\
    UC15 & 0.423 & 0.685 & 0.294 & 0.291 & 0.725 & 0.740 \\
    UC16 & 0.429 & 0.641 & 0.238 & 0.242 & 0.680 & 0.721 \\
    UC17 & 0.296 & 0.530 & 0.141 & 0.138 & 0.556 & 0.678 \\
    UC18 & 0.281 & 0.523 & 0.120 & 0.127 & 0.554 & 0.675 \\
    UC19 & 0.242 & 0.712 & 0.310 & 0.313 & 0.776 & 0.752 \\
    UC20 & 0.227 & 0.479 & 0.079 & 0.084 & 0.512 & 0.659 \\
    UC21 & 0.222 & 0.693 & 0.307 & 0.307 & 0.719 & 0.744 \\
    UC22 & 0.214 & 0.672 & 0.285 & 0.270 & 0.724 & 0.735 \\
    UC23 & 0.161 & 0.626 & 0.241 & 0.219 & 0.677 & 0.715 \\
    UC24 & 0.140 & 0.487 & 0.079 & 0.100 & 0.509 & 0.662 \\
    UC25 & 0.107 & 0.476 & 0.089 & 0.085 & 0.504 & 0.658 \\
    UC26 & 0.107 & 0.560 & 0.166 & 0.161 & 0.595 & 0.689 \\
    UC27 & 0.033 & 0.534 & 0.147 & 0.131 & 0.573 & 0.679 \\
    UC28 & 0.028 & 0.492 & 0.134 & 0.106 & 0.512 & 0.664 \\
    UC29 & 0.020 & 0.645 & 0.246 & 0.254 & 0.670 & 0.723 \\
    UC30 & 0.002 & 0.384 & 0.000 & 0.000 & 0.413 & 0.625 \\
    \hline
    \end{tabular}%
  \label{tab:data}%
\end{table}%

\begin{table}[H]
\centering
\caption{Results obtained for the \textbf{GeReSiD50} benchmark dataset by different methods in isolation}
\resizebox{!}{.35\paperheight}{%
\begin{tabular}{|c|c|c|c|c|c|c|}
\hline
\textbf{UC} & \textbf{Truth} & \textbf{Bert-Cos.} & \textbf{Bert-Man.} & \textbf{Bert-Euc.} & \textbf{Bert-Inn.} & \textbf{Bert-Ang.}\\
\hline
UC1 & 0.017 & 0.320 & 0.046 & 0.133 & 0.373 & 0.604 \\
UC2 & 0.021 & 0.275 & 0.054 & 0.109 & 0.316 & 0.589 \\
UC3 & 0.031 & 0.391 & 0.139 & 0.193 & 0.440 & 0.628 \\
UC4 & 0.050 & 0.450 & 0.160 & 0.220 & 0.525 & 0.649 \\
UC5 & 0.052 & 0.174 & 0.000 & 0.050 & 0.200 & 0.556 \\
UC6 & 0.058 & 0.544 & 0.238 & 0.300 & 0.616 & 0.683 \\
UC7 & 0.072 & 0.354 & 0.089 & 0.160 & 0.408 & 0.615 \\
UC8 & 0.081 & 0.563 & 0.260 & 0.310 & 0.646 & 0.690 \\
UC9 & 0.085 & 0.240 & 0.015 & 0.080 & 0.281 & 0.577 \\
UC10 & 0.094 & 0.233 & 0.025 & 0.088 & 0.267 & 0.575 \\
UC11 & 0.109 & 0.152 & 0.000 & 0.023 & 0.181 & 0.549 \\
UC12 & 0.124 & 0.377 & 0.098 & 0.164 & 0.446 & 0.623 \\
UC13 & 0.139 & 0.394 & 0.133 & 0.181 & 0.460 & 0.629 \\
UC14 & 0.149 & 0.477 & 0.163 & 0.228 & 0.571 & 0.658 \\
UC15 & 0.154 & 0.497 & 0.207 & 0.258 & 0.574 & 0.666 \\
UC16 & 0.161 & 0.683 & 0.374 & 0.428 & 0.742 & 0.739 \\
UC17 & 0.204 & 0.368 & 0.099 & 0.164 & 0.428 & 0.620 \\
UC18 & 0.210 & 0.606 & 0.299 & 0.354 & 0.677 & 0.707 \\
UC19 & 0.217 & 0.456 & 0.185 & 0.234 & 0.519 & 0.651 \\
UC20 & 0.235 & 0.366 & 0.121 & 0.176 & 0.414 & 0.619 \\
UC21 & 0.269 & 0.634 & 0.310 & 0.359 & 0.749 & 0.719 \\
UC22 & 0.273 & 0.319 & 0.095 & 0.139 & 0.365 & 0.603 \\
UC23 & 0.290 & 0.510 & 0.204 & 0.271 & 0.582 & 0.670 \\
UC24 & 0.328 & 0.603 & 0.279 & 0.339 & 0.700 & 0.706 \\
UC25 & 0.369 & 0.413 & 0.122 & 0.184 & 0.493 & 0.635 \\
UC26 & 0.389 & 0.506 & 0.200 & 0.256 & 0.597 & 0.669 \\
UC27 & 0.391 & 0.768 & 0.456 & 0.497 & 0.883 & 0.779 \\
UC28 & 0.399 & 0.676 & 0.356 & 0.404 & 0.782 & 0.736 \\
UC29 & 0.417 & 0.669 & 0.348 & 0.395 & 0.776 & 0.733 \\
UC30 & 0.438 & 0.501 & 0.197 & 0.255 & 0.587 & 0.667 \\
UC31 & 0.490 & 0.639 & 0.315 & 0.369 & 0.740 & 0.720 \\
UC32 & 0.514 & 0.427 & 0.136 & 0.206 & 0.495 & 0.640 \\
UC33 & 0.535 & 0.497 & 0.191 & 0.259 & 0.571 & 0.666 \\
UC34 & 0.557 & 0.492 & 0.174 & 0.236 & 0.594 & 0.664 \\
UC35 & 0.594 & 0.800 & 0.500 & 0.534 & 0.915 & 0.795 \\
UC36 & 0.611 & 0.561 & 0.243 & 0.309 & 0.641 & 0.689 \\
UC37 & 0.617 & 0.753 & 0.444 & 0.480 & 0.868 & 0.771 \\
UC38 & 0.621 & 0.713 & 0.400 & 0.442 & 0.815 & 0.753 \\
UC39 & 0.645 & 0.532 & 0.230 & 0.284 & 0.614 & 0.679 \\
UC40 & 0.650 & 0.665 & 0.354 & 0.400 & 0.750 & 0.731 \\
UC41 & 0.668 & 0.574 & 0.256 & 0.317 & 0.665 & 0.695 \\
UC42 & 0.748 & 0.920 & 0.682 & 0.706 & 1.053 & 0.872 \\
UC43 & 0.762 & 0.704 & 0.385 & 0.426 & 0.826 & 0.749 \\
UC44 & 0.764 & 0.631 & 0.330 & 0.372 & 0.710 & 0.717 \\
UC45 & 0.764 & 0.726 & 0.399 & 0.449 & 0.848 & 0.759 \\
UC46 & 0.769 & 0.658 & 0.333 & 0.391 & 0.751 & 0.729 \\
UC47 & 0.781 & 0.572 & 0.248 & 0.312 & 0.666 & 0.694 \\
UC48 & 0.811 & 0.651 & 0.322 & 0.382 & 0.750 & 0.726 \\
UC49 & 0.873 & 0.751 & 0.425 & 0.475 & 0.876 & 0.770 \\
UC50 & 0.904 & 0.866 & 0.588 & 0.617 & 1.000 & 0.834 \\
\hline
\end{tabular}}
  \label{tab:data2}
\end{table}

In first instance, we have compared GE with a simple BERT-based ensemble that averages cosine, Euclidean, and Manhattan distances. Our approach is able to consistently get better results that the BERT-based ensemble, confirming that our evolutionary approach outperforms naive similarity aggregation. However, it is important to remark that the outcomes of our reported experiments are based on 30 independent runs, owing to the inherent non-deterministic characteristics of the methods. Therefore, we aim to report below a snapshot of the values achieved using much stronger baselines.

\subsection {Assessing Semantic Similarity in a General-purpose Context}
Figure \ref{fig:r1} shows the results for two evaluation criteria, PCC and SRCC, over the \textbf{MC30} benchmark dataset. The x-axis represents different strategies used for evaluation. At the same time, Linear Regression (LR) is the baseline, as discussed earlier. A dotted horizontal line represents it. 

The state-of-the-art genetic ensembles are Tree-based Genetic Programming (TGP) \cite{key-koza}, Linear Genetic Programming (LGP) \cite{key-Brameier}, and Cartesian Genetic Programming (CGP) \cite{key-Miller20} precisely as in \cite{key-martinez-ijseke}. GE is the approach proposed in this work, and GE-i is the interpretable variant of GE discussed earlier. It is important to note that all the ensembles are trained on the same training dataset to facilitate the fairness of the comparisons.

In the first subplot (a), the LGP achieves relatively high performance compared to the other methods. The boxplot shows the distribution of PCC values obtained from 30 experimental runs. The box represents the interquartile range (IQR), where the central box spans from the lower quartile (Q1) to the upper quartile (Q3). The line within the box corresponds to the median value. The whiskers extend to the minimum and maximum values.

In the second subplot (b), the GE method (first blue boxplot) demonstrates the best performance regarding SRCC. The boxplot characteristics are the same as in the previous subplot but now represent the distribution of SRCC values.

Both subplots suggest that the LGP outperforms the other evaluated methods regarding PCC, and GE is superior regarding SRCC on the \textbf{MC30} benchmark dataset. GE-i, although interpretable, achieves the worst performance.

\begin{figure}[h]
\centering
\begin{tikzpicture}[scale=1.00]
\begin{axis}[
boxplot/draw direction=y,
xtick={1,2,3,4,5},
xticklabels={TGP, LGP, CGP, \textbf{GE}, \textbf{GE-i}},
]
\draw[dotted] (axis cs:0,0.757) -- (axis cs:6,0.757);

\addplot[color=black]  [boxplot prepared={draw position=1,
lower whisker=0.752, lower quartile=0.752,
median=0.753, upper quartile=0.753,
upper whisker=0.754},
] coordinates {};
\addplot[color=black]  [boxplot prepared={draw position=2,
lower whisker=0.796, lower quartile=0.83,
median=0.84, upper quartile=0.85,
upper whisker=0.86},
] coordinates {};
\addplot[color=black]  [boxplot prepared={draw position=3,
lower whisker=0.752, lower quartile=0.755,
median=0.764, upper quartile=0.766,
upper whisker=0.768},
] coordinates {};
\addplot[color=blue]  [boxplot prepared={draw position=4,
lower whisker=0.744, lower quartile=0.755,
median=0.786, upper quartile=0.797,
upper whisker=0.826},
] coordinates {};
\addplot[color=blue]  [boxplot prepared={draw position=5,
lower whisker=0.750, lower quartile=0.750,
median=0.751, upper quartile=0.751,
upper whisker=0.751},
] coordinates {};
\end{axis}
\end{tikzpicture}
\begin{tikzpicture}[scale=1.00]
\begin{axis}[
boxplot/draw direction=y,
xtick={1,2,3,4,5},
xticklabels={TGP, LGP, CGP, \textbf{GE}, \textbf{GE-i}},
]

\draw[dotted] (axis cs:0,0.770) -- (axis cs:6,0.770);

\addplot[color=black]  [boxplot prepared={draw position=1,
lower whisker=0.752, lower quartile=0.753,
median=0.754, upper quartile=0.754,
upper whisker=0.755},
] coordinates {};
\addplot[color=black]  [boxplot prepared={draw position=2,
lower whisker=0.783, lower quartile=0.797,
median=0.820, upper quartile=0.835,
upper whisker=0.855},
] coordinates {};
\addplot[color=black]  [boxplot prepared={draw position=3,
lower whisker=0.752, lower quartile=0.757,
median=0.763, upper quartile=0.767,
upper whisker=0.767},
] coordinates {};
\addplot[color=blue]  [boxplot prepared={draw position=4,
lower whisker=0.799, lower quartile=0.825,
median=0.856, upper quartile=0.867,
upper whisker=0.88},
] coordinates {};
\addplot[color=blue]  [boxplot prepared={draw position=5,
lower whisker=0.717, lower quartile=0.717,
median=0.718, upper quartile=0.718,
upper whisker=0.718},
] coordinates {};
\end{axis}
\end{tikzpicture}
\caption{Results for the a) \textbf{PCC} and b) \textbf{SRCC} over the \textbf{MC30} benchmark dataset}
\label{fig:r1}
\end{figure}

As a matter of curiosity, we can see in Example \ref{ex:A} the code generated for both PCC ans SRCC over the MC30 dataset. This given source code is represented in Python and uses the Numpy library, which supports mathematical operations on arrays and matrices. The result is computed using various mathematical functions and operators. The reason is that we are using the FG seen in Example \ref{ex:01}. It is important to note that the expressions within parentheses are evaluated and combined using the specified operators.

\small
\begin{example}[label=ex:A]
\begin{tcolorbox}[colback=black!10!white,colframe=black!70!black,title=Ensemble optimized for PCC over MC30,colbacktitle=black!50!white]
\begin{verbatim}
import numpy as np

result = (
    BERT-Euc - BERT-Inn + pdiv(BERT-Euc, np.sin(BERT-Ang)) - BERT-Cos +
    np.exp(psqrt(pdiv(np.tanh(BERT-Man), BERT-Ang))) + BERT-Euc +
    psqrt(pdiv(BERT-Cos, np.sin(pdiv(BERT-Inn, pdiv(np.sin(BERT-Ang), 
		BERT-Inn) * BERT-Man - BERT-Euc) * pdiv(BERT-Man, BERT-Inn))))
) / (BERT-Inn - pdiv(71.24, BERT-Cos * plog(76.12)))
\end{verbatim}
\end{tcolorbox}
\begin{tcolorbox}[colback=black!10!white,colframe=black!70!black,title=Ensemble optimized for SRCC over MC30,colbacktitle=black!50!white]
\begin{verbatim}
import numpy as np

result = (
    BERT-Euc - BERT-Inn + pdiv(BERT-Euc, np.sin(BERT-Ang)) - BERT-Cos +
    np.exp(psqrt(pdiv(np.tanh(BERT-Man), BERT-Ang))) + BERT-Euc +
    psqrt(pdiv(BERT-Cos, np.sin(pdiv(BERT-Inn, pdiv(np.sin(BERT-Ang), 
		BERT-Inn) * BERT-Man - BERT-Euc) * pdiv(BERT-Man, BERT-Inn))))
) / (BERT-Inn - pdiv(71.24, BERT-Cos * plog(76.12)))
\end{verbatim}
\end{tcolorbox}
\end{example}
\normalsize

We also show the changes over time in important variables during the GE process. Figure \ref{fig:metricsaa} shows the progression of these parameters. Specifically, we focus on four key parameters: Average Fitness, Average Genome Length, Average Tree Nodes, and Best Fitness.

The \textbf{Average Fitness} provides insights into the overall performance of the evolving population. It reflects the average fitness value of individuals in each generation, indicating the progress achieved by the GE strategy. A steady increase in Average Fitness over generations shows that the evolutionary process is successfully refining ensemble configurations to better approximate human judgments. However, plateaus or sharp fluctuations may indicate premature convergence or excessive randomness in the search process.

The \textbf{Average Genome Length} tracks the average length of individual genomes within the population at different training stages. The goal of monitoring this variable is to understand how the complexity of GE-generated solutions changes over time. This is crucial because longer genomes indicate more complex aggregation formulas, which may lead to overfitting on the training set. 

The \textbf{Average Tree Nodes} measures the average number of nodes in the evolved solutions. It offers valuable information about the complexity of evolved ensembles, shedding light on the strategy's search for space exploration. This metric helps determine whether the evolved formulas are becoming too complex for practical interpretability. 

Lastly, the \textbf{Best Fitness} represents the fitness value of the best individual in each generation. Observing this variable helps to assess the progress in finding optimal solutions as training is performed. A steady increase in Best Fitness indicates that GE is finding progressively better solutions, but please remember that these values are for the training phase, and then it remains to test the generated ensemble on previously unseen data.

Analyzing the evolution of these variables allows us to obtain insights into how they contribute to PCC optimization and interact during the GE process over the MC30 benchmark dataset. This analysis provides a valuable view into the behavior of GE and the performance of the approach.

\begin{figure}[h]
	\centering
\begin{subfigure}{0.45\textwidth}
\begin{tikzpicture}
\begin{axis}[
    xlabel={Generations},
    ylabel={Average Fitness},
    xmin=0, xmax=200,
    ymin=0.62, ymax=0.9,
    xtick={0,50,100,150,200},
    ytick={0.65,0.70,0.75,0.80,0.85,0.9},
    legend pos=north east,
    ymajorgrids=true,
    grid style=dashed,
]

\addplot[color=blue, mark=none] table[x=a, y=b] {data00.txt};
\end{axis}
\end{tikzpicture}
\end{subfigure}
\hfill
\begin{subfigure}{0.45\textwidth}
\begin{tikzpicture}
\begin{axis}[
    xlabel={Generations},
    ylabel={Average Genome Length},
    xmin=0, xmax=200,
    ymin=0, ymax=180,
    xtick={0,50,100,150,200},
    ytick={0,20,...,180},
		legend pos=north east,
    ymajorgrids=true,
    grid style=dashed,
]

\addplot[red,line width=1pt] table[x=a, y=b] {data01.txt};
\end{axis}
\end{tikzpicture}
\end{subfigure}
    
    \vspace{0.4cm}
    
\begin{subfigure}{0.45\textwidth}
\begin{tikzpicture}
\begin{axis}[
    xlabel={Generations},
    ylabel={Average Tree Nodes},
    xmin=0, xmax=200,
    ymin=0, ymax=110,
    xtick={0,50,100,150,200},
    ytick={10, 20, 30, 40, 50, 60, 70, 80, 90, 100, 110},
    legend pos=north east,
    ymajorgrids=true,
    grid style=dashed,
]
\addplot[color=black, mark=none] table[x=a, y=b] {data02.txt};
\end{axis}
\end{tikzpicture}
\end{subfigure}
    \hfill
    \begin{subfigure}{0.45\textwidth}
		 \begin{tikzpicture}
\begin{axis}[
    xlabel={Generations},
    ylabel={Best Fitness},
    xmin=0, xmax=200,
    ymin=0.85, ymax=0.93,
    xtick={0,50,100,150,200},
    ytick={0.85,0.86,0.87,0.88,0.89,0.90,0.91,0.92,0.93},
    legend pos=north east,
    ymajorgrids=true,
    grid style=dashed,
]
\addplot[color=purple, mark=none] table[x=a, y=b] {data03.txt};
\end{axis}
\end{tikzpicture}  
			
			\end{subfigure}
    
    \caption{Evolution of different variables during the ensemble learning process for \textbf{PCC} over the \textbf{MC30} benchmark dataset}
    \label{fig:metricsaa}
\end{figure}

Figure \ref{fig:metricsbb} reports a comprehensive visualization concerning the progressive evolution of the aforementioned important variables when optimizing SRCC over the \textbf{MC30} benchmark dataset.

\begin{figure}[h]
	\centering
\begin{subfigure}{0.45\textwidth}
\begin{tikzpicture}
\begin{axis}[
    xlabel={Generations},
    ylabel={Average Fitness},
    xmin=0, xmax=200,
    ymin=0.57, ymax=0.9,
    xtick={0,50,100,150,200},
    ytick={0,6,0.65,0.70,0.75,0.80,0.85,0.9},
    legend pos=north east,
    ymajorgrids=true,
    grid style=dashed,
]
\addplot[color=blue, mark=none] table[x=a, y=b] {data10.txt};
\end{axis}
\end{tikzpicture}
\end{subfigure}
\hfill
\begin{subfigure}{0.45\textwidth}
\begin{tikzpicture}
\begin{axis}[
    xlabel={Generations},
    ylabel={Average Genome Length},
    xmin=0, xmax=200,
    ymin=0, ymax=350,
    xtick={0,50,100,150,200},
    ytick={0,50,150,250,350},
		legend pos=north east,
    ymajorgrids=true,
    grid style=dashed,
]

\addplot[red,line width=1pt] table[x=a, y=b] {data11.txt};
\end{axis}
\end{tikzpicture}
\end{subfigure}
    
    \vspace{0.4cm}
    
    \begin{subfigure}{0.45\textwidth}

\begin{tikzpicture}
\begin{axis}[
    xlabel={Generations},
    ylabel={Average Tree Nodes},
    xmin=0, xmax=200,
    ymin=0, ymax=80,
    xtick={0,50,100,150,200},
    ytick={0, 20, 40, 60, 80},
    legend pos=north east,
    ymajorgrids=true,
    grid style=dashed,
]
\addplot[color=black, mark=none] table[x=a, y=b] {data12.txt};
\end{axis}
\end{tikzpicture}
\end{subfigure}
    \hfill
    \begin{subfigure}{0.45\textwidth}
		 \begin{tikzpicture}
\begin{axis}[
    xlabel={Generations},
    ylabel={Best Fitness},
    xmin=0, xmax=200,
    ymin=0.7, ymax=0.95,
    xtick={0,50,100,150,200},
    ytick={0.7,0.75,0.8,0.85,0.90,0.95},
    legend pos=north east,
    ymajorgrids=true,
    grid style=dashed,
]
\addplot[color=purple, mark=none] table[x=a, y=b] {data13.txt};
\end{axis}
\end{tikzpicture}  
			
			\end{subfigure}
    
    \caption{Evolution of different variables during the ensemble learning process for \textbf{SRCC} over the \textbf{MC30} benchmark dataset}
    \label{fig:metricsbb}
\end{figure}

\subsection {Assessing Semantic Similarity in a Domain-Specific Context}
Figure \ref{fig:r2} shows the results for both PCC and SRCC over the \textbf{GeReSiD50} benchmark dataset. As in the previous case, the x-axis represents different strategies used for evaluation. Linear Regression (LR) is again the baseline, as discussed earlier, and is represented by a dotted horizontal line. The state-of-the-art genetic ensembles are again TGP \cite{key-koza}, LGP \cite{key-Brameier}, and CGP \cite{key-Miller20}. GE is again the approach proposed in this work, and GE-i is the interpretable variant of GE, precisely as we discussed in the previous case.

In the first subplot (a), the LGP achieves relatively high performance compared to the other methods. The boxplot shows the distribution of PCC values obtained from 30 experimental runs. The box again represents the IQR, where the central box spans from the lower to the upper quartile. The line within the box corresponds to the median value. 

In the second subplot (b), the GE method achieves the best performance regarding SRCC. The boxplot characteristics are the same as in the previous subplot but now represent the distribution of SRCC values. It is possible to see that, as with the general purpose use case, both subplots suggest again that the LGP outperforms the other evaluated methods regarding PCC, and GE is superior regarding SRCC on the \textbf{GeReSiD50} benchmark dataset. GE-i, although interpretable, achieves the worst performance once again.

\begin{figure}[h]
\centering
\begin{tikzpicture}[scale=1.00]
\begin{axis}[
boxplot/draw direction=y,
xtick={1,2,3,4,5},
xticklabels={TGP, LGP, CGP, \textbf{GE}, \textbf{GE-i}},
]
\draw[dotted] (axis cs:0,0.736) -- (axis cs:6,0.736);

\addplot[color=black]  [boxplot prepared={draw position=1,
lower whisker=0.734, lower quartile=0.735,
median=0.735, upper quartile=0.736,
upper whisker=0.737},
] coordinates {};
\addplot[color=black]  [boxplot prepared={draw position=2,
lower whisker=0.737, lower quartile=0.741,
median=0.756, upper quartile=0.763,
upper whisker=0.776},
] coordinates {};
\addplot[color=black]  [boxplot prepared={draw position=3,
lower whisker=0.724, lower quartile=0.736,
median=0.737, upper quartile=0.741,
upper whisker=0.743},
] coordinates {};
\addplot[color=blue]  [boxplot prepared={draw position=4,
lower whisker=0.738, lower quartile=0.741,
median=0.743, upper quartile=0.746,
upper whisker=0.749},
] coordinates {};
\addplot[color=blue]  [boxplot prepared={draw position=5,
lower whisker=0.734, lower quartile=0.734,
median=0.735, upper quartile=0.735,
upper whisker=0.736},
] coordinates {};
\end{axis}
\end{tikzpicture}
\begin{tikzpicture}[scale=1.00]
\begin{axis}[
boxplot/draw direction=y,
xtick={1,2,3,4,5},
xticklabels={TGP, LGP, CGP, \textbf{GE}, \textbf{GE-i}},
]

\draw[dotted] (axis cs:0,0.744) -- (axis cs:6,0.744);

\addplot[color=black]  [boxplot prepared={draw position=1,
lower whisker=0.738, lower quartile=0.738,
median=0.74, upper quartile=0.74,
upper whisker=0.741},
] coordinates {};
\addplot[color=black]  [boxplot prepared={draw position=2,
lower whisker=0.733, lower quartile=0.747,
median=0.756, upper quartile=0.765,
upper whisker=0.771},
] coordinates {};
\addplot[color=black]  [boxplot prepared={draw position=3,
lower whisker=0.737, lower quartile=0.739,
median=0.742, upper quartile=0.745,
upper whisker=0.747},
] coordinates {};
\addplot[color=blue]  [boxplot prepared={draw position=4,
lower whisker=0.774, lower quartile=0.777,
median=0.779, upper quartile=0.78,
upper whisker=0.781},
] coordinates {};
\addplot[color=blue]  [boxplot prepared={draw position=5,
lower whisker=0.739, lower quartile=0.739,
median=0.739, upper quartile=0.74,
upper whisker=0.74},
] coordinates {};
\end{axis}
\end{tikzpicture}
\caption{Results for the a) \textbf{PCC} and b) \textbf{SRCC} over the \textbf{GeReSiD50} benchmark dataset}
\label{fig:r2}
\end{figure}

As a matter of curiosity, we provide the generated Python source code in Example \ref{ex:B}. It is an ensemble optimized for PCC over MC30 that consists of two functions: $my\_pearson(x, y)$ and $p()$. The $my\_pearson(x, y)$ function calculates the PCC between two arrays, while the $p()$ function aims to maximize through an algebraic formula that needs to be learned. In order to do that, the code reads data from training and validation CSV files, extracts relevant columns, and performs calculations to generate a new column with the expression to be learned. The PCC coefficient between the $response$ column and the new column is then computed, which serves as the goal (PCC over unseen data) to be maximized.

\small
\begin{example}[label=ex:B]
\begin{tcolorbox}[colback=black!10!white,colframe=black!70!black,title=Ensemble optimized for PCC over MC30,colbacktitle=black!50!white]
\begin{verbatim}
  import pandas as pd
  import numpy as np
	
  def my_pearson(x, y):
  	 return np.abs(np.corrcoef(x, y)[0,1])

  def p():

     df = pd.read_csv('c:/mc-training.txt')
     df2 = pd.read_csv('c:/mc-validation.txt')
	
     x, x0, x1, x2, x3, x4 = df['response'].to_numpy(),\ 
      df['x0'].to_numpy(), df['x1'].to_numpy(), df['x2'].to_numpy(), \
      df['x3'].to_numpy(), df['x4'].to_numpy()
  
	   y, y0, y1, y2, y3, y4 = df2['response'].to_numpy(),\ 
      df2['y0'].to_numpy(), df2['y1'].to_numpy(), df2['y2'].to_numpy(), \
      df2['y3'].to_numpy(), df2['y4'].to_numpy()
  
     aux = 'np.sin(x2)'
     aux2 = aux.replace('x','y')
     df2['new'] = eval(aux2)
	
     return my_pearson(y, df2['new'].to_numpy())
\end{verbatim}
\end{tcolorbox}
\end{example}
\normalsize

We also provide the generated code for SRCC in Example \ref{ex:C}. It is an ensemble optimized for SRCC over MC30 that implements two functions, $my\_spearman(x, y)$ and $p()$, to maximize SRCC. The $my\_spearman(x, y)$ function calculates the SRCC between two arrays. The $p()$ function loads training and validation datasets, extracts relevant columns, and performs calculations on the data. 

The ensemble defines an auxiliary expression involving variables, replaces one set of variables with another, evaluates the expression, and assigns the results to a new column in the validation dataset. Finally, the SRCC is computed between the $response$ and newly created columns. The objective is maximizing the value returned by $p()$, representing the SRCC over unseen data.

\small
\begin{example}[label=ex:C]
\begin{tcolorbox}[colback=black!10!white,colframe=black!70!black,title=Ensemble optimized for SRCC over MC30,colbacktitle=black!50!white]
\begin{verbatim}
  import pandas as pd
  import numpy as np
  from scipy.stats import spearmanr
	
  def my_spearman(x, y):
  	 return np.abs(spearmanr(x, y)[0])

  def p():

     df = pd.read_csv('c:/geresid-training.txt')
     df2 = pd.read_csv('c:/geresid-validation.txt')
	
     x, x0, x1, x2, x3, x4 = df['response'].to_numpy(),\ 
      df['x0'].to_numpy(), df['x1'].to_numpy(), df['x2'].to_numpy(), \
      df['x3'].to_numpy(), df['x4'].to_numpy()
  
	   y, y0, y1, y2, y3, y4 = df2['response'].to_numpy(),\ 
      df2['y0'].to_numpy(), df2['y1'].to_numpy(), df2['y2'].to_numpy(), \
      df2['y3'].to_numpy(), df2['y4'].to_numpy()
  
     aux = 'x3 * x3 * x4'
     aux2 = aux.replace('x','y')
     df2['new'] = eval(aux2)
	
     return my_spearman(y, df2['new'].to_numpy())
\end{verbatim}
\end{tcolorbox}
\end{example}
\normalsize

Once again, examining Figure \ref{fig:metricscc} deepens our understanding of the progressive evolution of critical variables in generating the ensemble using \textbf{PCC} over the \textbf{GeReSiD50} benchmark dataset. This analysis sheds light on the optimization process's behavior and the interplay between critical parameters.

\begin{figure}[h]
	\centering
\begin{subfigure}{0.45\textwidth}
\begin{tikzpicture}
\begin{axis}[
    xlabel={Generations},
    ylabel={Average Fitness},
    xmin=0, xmax=200,
    ymin=0.62, ymax=0.9,
    xtick={0,50,100,150,200},
    ytick={0.65,0.70,0.75,0.80,0.85,0.9},
    legend pos=north east,
    ymajorgrids=true,
    grid style=dashed,
]

\addplot[color=blue, mark=none] table[x=a, y=b] {data20.txt};
\end{axis}
\end{tikzpicture}
\end{subfigure}
\hfill
\begin{subfigure}{0.45\textwidth}
\begin{tikzpicture}
\begin{axis}[
    xlabel={Generations},
    ylabel={Average Genome Length},
    xmin=0, xmax=200,
    ymin=0, ymax=400,
    xtick={0,50,100,150,200},
    ytick={0,100,200,300,400},
		legend pos=north east,
    ymajorgrids=true,
    grid style=dashed,
]

\addplot[red,line width=1pt] table[x=a, y=b] {data21.txt};
\end{axis}
\end{tikzpicture}
\end{subfigure}
    
    \vspace{0.4cm}
    
\begin{subfigure}{0.45\textwidth}
\begin{tikzpicture}
\begin{axis}[
    xlabel={Generations},
    ylabel={Average Tree Nodes},
    xmin=0, xmax=200,
    ymin=0, ymax=100,
    xtick={0,50,100,150,200},
    ytick={0, 20, 40, 60, 80, 100},
    legend pos=north east,
    ymajorgrids=true,
    grid style=dashed,
]
\addplot[color=black, mark=none] table[x=a, y=b] {data22.txt};
\end{axis}
\end{tikzpicture}
\end{subfigure}
    \hfill
    \begin{subfigure}{0.45\textwidth}
	\begin{tikzpicture}
	\begin{axis}[
    xlabel={Generations},
    ylabel={Best Fitness},
    xmin=0, xmax=200,
    ymin=0.85, ymax=0.9,
    xtick={0,50,100,150,200},
    ytick={0.85,0.86,0.87,0.88,0.89,0.90},
    legend pos=north east,
    ymajorgrids=true,
    grid style=dashed,
]
\addplot[color=purple, mark=none] table[x=a, y=b] {data23.txt};
\end{axis}
\end{tikzpicture}  	
\end{subfigure}
   
    \caption{Evolution of key variables during the ensemble learning process for \textbf{PCC} over the \textbf{GeReSiD50} dataset}
    \label{fig:metricscc}
\end{figure}

At the same time, and once again, Figure \ref{fig:metricsdd} shows us the progressive evolution of these critical variables. However, this time is intended to understand better the process of optimizing \textbf{SRCC} over the \textbf{GeReSiD50} benchmark dataset.

\begin{figure}[h]
	\centering
\begin{subfigure}{0.45\textwidth}
\begin{tikzpicture}
\begin{axis}[
    xlabel={Generations},
    ylabel={Average Fitness},
    xmin=0, xmax=200,
    ymin=0.62, ymax=0.95,
    xtick={0,50,100,150,200},
    ytick={0.65,0.70,0.75,0.80,0.85,0.9,0.95},
    legend pos=north east,
    ymajorgrids=true,
    grid style=dashed,
]

\addplot[color=blue, mark=none] table[x=a, y=b]{data30.txt};
\end{axis}
\end{tikzpicture}
\end{subfigure}
\hfill
\begin{subfigure}{0.45\textwidth}
\begin{tikzpicture}
\begin{axis}[
    xlabel={Generations},
    ylabel={Average Genome Length},
    xmin=0, xmax=200,
    ymin=0, ymax=180,
    xtick={0,50,100,150,200},
    ytick={0,20,...,180},
		legend pos=north east,
    ymajorgrids=true,
    grid style=dashed,
]

\addplot[red,line width=1pt] table[x=a, y=b]{data31.txt};
\end{axis}
\end{tikzpicture}
\end{subfigure}
    
    \vspace{0.4cm}
    
    \begin{subfigure}{0.45\textwidth}

\begin{tikzpicture}
\begin{axis}[
    xlabel={Generations},
    ylabel={Average Tree Nodes},
    xmin=0, xmax=200,
    ymin=0, ymax=100,
    xtick={0,50,100,150,200},
    ytick={10, 20, 30, 40, 50, 60, 70, 80, 90, 100},
    legend pos=north east,
    ymajorgrids=true,
    grid style=dashed,
]
\addplot[color=black, mark=none] table[x=a, y=b]{data32.txt};
\end{axis}
\end{tikzpicture}
\end{subfigure}
    \hfill
    \begin{subfigure}{0.45\textwidth}
		 \begin{tikzpicture}
\begin{axis}[
    xlabel={Generations},
    ylabel={Best Fitness},
    xmin=0, xmax=200,
    ymin=0.85, ymax=0.93,
    xtick={0,50,100,150,200},
    ytick={0.85,0.86,0.87,0.88,0.89,0.90,0.91,0.92,0.93},
    legend pos=north east,
    ymajorgrids=true,
    grid style=dashed,
]
\addplot[color=purple, mark=none] table[x=a, y=b]{data33.txt};
\end{axis}
\end{tikzpicture}  
\end{subfigure}
    
    \caption{Evolution of key variables during the ensemble learning process for \textbf{SRCC} over the \textbf{GeReSiD50} dataset}
    \label{fig:metricsdd}
\end{figure}

\subsection{Assessing Semantic Similarity with a Large Dataset}
Figure \ref{fig:r3} presents the results for two evaluation criteria, PCC and SRCC, on the \textbf{WS353} benchmark dataset. The x-axis indicates the various evaluation strategies, with LR as the baseline, represented by a dotted horizontal line.

The state-of-the-art genetic ensembles included are TGP \cite{key-koza}, LGP \cite{key-Brameier}, and CGP \cite{key-Miller20}, as detailed in \cite{key-martinez-ijseke}. GE refers to the proposed approach in this work, while GE-i denotes its interpretable variant discussed earlier. All ensembles were again trained on the same training dataset to ensure fair comparisons.

\begin{figure}[h]
\centering
\begin{tikzpicture}[scale=1.00]
\begin{axis}[
boxplot/draw direction=y,
xtick={1,2,3,4,5},
xticklabels={TGP, LGP, CGP, \textbf{GE}, \textbf{GE-i}},
]
\draw[dotted] (axis cs:0,0.28) -- (axis cs:6,0.28);

\addplot[color=black]  [boxplot prepared={draw position=1,
lower whisker=0.800, lower quartile=0.810,
median=0.811, upper quartile=0.811,
upper whisker=0.811},
] coordinates {};
\addplot[color=black]  [boxplot prepared={draw position=2,
lower whisker=0.760, lower quartile=0.761,
median=0.811, upper quartile=0.817,
upper whisker=0.817},
] coordinates {};
\addplot[color=black]  [boxplot prepared={draw position=3,
lower whisker=0.811, lower quartile=0.811,
median=0.811, upper quartile=0.811,
upper whisker=0.811},
] coordinates {};
\addplot[color=blue]  [boxplot prepared={draw position=4,
lower whisker=0.801, lower quartile=0.820,
median=0.827, upper quartile=0.827,
upper whisker=0.827},
] coordinates {};
\addplot[color=blue]  [boxplot prepared={draw position=5,
lower whisker=0.801, lower quartile=0.810,
median=0.811, upper quartile=0.812,
upper whisker=0.812},
] coordinates {};
\end{axis}
\end{tikzpicture}
\begin{tikzpicture}[scale=1.00]
\begin{axis}[
boxplot/draw direction=y,
xtick={1,2,3,4,5},
xticklabels={TGP, LGP, CGP, \textbf{GE}, \textbf{GE-i}},
]

\draw[dotted] (axis cs:0,0.48) -- (axis cs:6,0.48);

\addplot[color=black]  [boxplot prepared={draw position=1,
lower whisker=0.802, lower quartile=0.810,
median=0.812, upper quartile=0.812,
upper whisker=0.812},
] coordinates {};
\addplot[color=black]  [boxplot prepared={draw position=2,
lower whisker=0.767, lower quartile=0.795,
median=0.812, upper quartile=0.82,
upper whisker=0.833},
] coordinates {};
\addplot[color=black]  [boxplot prepared={draw position=3,
lower whisker=0.81, lower quartile=0.811,
median=0.811, upper quartile=0.811,
upper whisker=0.812},
] coordinates {};
\addplot[color=blue]  [boxplot prepared={draw position=4,
lower whisker=0.802, lower quartile=0.811,
median=0.817, upper quartile=0.817,
upper whisker=0.817},
] coordinates {};
\addplot[color=blue]  [boxplot prepared={draw position=5,
lower whisker=0.801, lower quartile=0.802,
median=0.804, upper quartile=0.807,
upper whisker=0.811},
] coordinates {};
\end{axis}
\end{tikzpicture}
\caption{Results for the a) \textbf{PCC} and b) \textbf{SRCC} over the \textbf{WS353} benchmark dataset}
\label{fig:r3}
\end{figure}

Simultaneously, the analysis of Figure \ref{fig:metricsee} provides a detailed view of the progressive evolution of these critical variables. This time, the focus is on gaining deeper insights into the process of optimizing \textbf{PCC} on the \textbf{WS353} benchmark dataset.

\begin{figure}[h]
	\centering
\begin{subfigure}{0.45\textwidth}
\begin{tikzpicture}
\begin{axis}[
    xlabel={Generations},
    ylabel={Average Fitness},
    xmin=0, xmax=200,
    ymin=0.62, ymax=0.95,
    xtick={0,50,100,150,200},
    ytick={0.65,0.70,0.75,0.80,0.85,0.9,0.95},
    legend pos=north east,
    ymajorgrids=true,
    grid style=dashed,
]

\addplot[color=blue, mark=none] table[x=a, y=b]{data40.txt};
\end{axis}
\end{tikzpicture}
\end{subfigure}
\hfill
\begin{subfigure}{0.45\textwidth}
\begin{tikzpicture}
\begin{axis}[
    xlabel={Generations},
    ylabel={Average Genome Length},
    xmin=0, xmax=200,
    ymin=0, ymax=180,
    xtick={0,50,100,150,200},
    ytick={0,20,...,180},
		legend pos=north east,
    ymajorgrids=true,
    grid style=dashed,
]

\addplot[red,line width=1pt] table[x=a, y=b]{data41.txt};
\end{axis}
\end{tikzpicture}
\end{subfigure}
    
    \vspace{0.4cm}
    
\begin{subfigure}{0.45\textwidth}
\begin{tikzpicture}
\begin{axis}[
    xlabel={Generations},
    ylabel={Average Tree Nodes},
    xmin=0, xmax=200,
    ymin=0, ymax=100,
    xtick={0,50,100,150,200},
    ytick={10, 20, 30, 40, 50, 60, 70, 80, 90, 100},
    legend pos=north east,
    ymajorgrids=true,
    grid style=dashed,
]
\addplot[color=black, mark=none] table[x=a, y=b]{data42.txt};
\end{axis}
\end{tikzpicture}
\end{subfigure}
    \hfill
\begin{subfigure}{0.45\textwidth}
\begin{tikzpicture}
\begin{axis}[
    xlabel={Generations},
    ylabel={Best Fitness},
    xmin=0, xmax=200,
    ymin=0.78, ymax=0.86,
    xtick={0,50,100,150,200},
    ytick={0.78,0.79,0.80,0.81,0.82,0.83,0.84,0.85,0.86},
    legend pos=north east,
    ymajorgrids=true,
    grid style=dashed,
]
\addplot[color=purple, mark=none] table[x=a, y=b]{data43.txt};
\end{axis}
\end{tikzpicture}  
\end{subfigure}
\caption{Evolution of key variables during the ensemble learning process for \textbf{PCC} over the \textbf{WS353} dataset}
\label{fig:metricsee}
\end{figure}

Figure \ref{fig:metricsff} offers a detailed view of the progressive evolution of key variables, with a focus on the optimization of \textbf{SRCC} on the \textbf{WS353} benchmark dataset for deeper insights into the process.

\begin{figure}[h]
	\centering
\begin{subfigure}{0.45\textwidth}
\begin{tikzpicture}
\begin{axis}[
    xlabel={Generations},
    ylabel={Average Fitness},
    xmin=0, xmax=200,
    ymin=0.62, ymax=0.95,
    xtick={0,50,100,150,200},
    ytick={0.65,0.70,0.75,0.80,0.85,0.9,0.95},
    legend pos=north east,
    ymajorgrids=true,
    grid style=dashed,
]

\addplot[color=blue, mark=none] table[x=a, y=b]{data50.txt};
\end{axis}
\end{tikzpicture}
\end{subfigure}
\hfill
\begin{subfigure}{0.45\textwidth}
\begin{tikzpicture}
\begin{axis}[
    xlabel={Generations},
    ylabel={Average Genome Length},
    xmin=0, xmax=200,
    ymin=0, ymax=180,
    xtick={0,50,100,150,200},
    ytick={0,20,...,180},
		legend pos=north east,
    ymajorgrids=true,
    grid style=dashed,
]

\addplot[red,line width=1pt] table[x=a, y=b]{data51.txt};
\end{axis}
\end{tikzpicture}
\end{subfigure}
    
    \vspace{0.4cm}
    
\begin{subfigure}{0.45\textwidth}
\begin{tikzpicture}
\begin{axis}[
    xlabel={Generations},
    ylabel={Average Tree Nodes},
    xmin=0, xmax=200,
    ymin=0, ymax=100,
    xtick={0,50,100,150,200},
    ytick={10, 20, 30, 40, 50, 60, 70, 80, 90, 100},
    legend pos=north east,
    ymajorgrids=true,
    grid style=dashed,
]
\addplot[color=black, mark=none] table[x=a, y=b]{data52.txt};
\end{axis}
\end{tikzpicture}
\end{subfigure}
    \hfill
\begin{subfigure}{0.45\textwidth}
\begin{tikzpicture}
\begin{axis}[
    xlabel={Generations},
    ylabel={Best Fitness},
    xmin=0, xmax=200,
    ymin=0.78, ymax=0.86,
    xtick={0,50,100,150,200},
    ytick={0.78,0.79,0.80,0.81,0.82,0.83,0.84,0.85,0.86},
    legend pos=north east,
    ymajorgrids=true,
    grid style=dashed,
]
\addplot[color=purple, mark=none] table[x=a, y=b]{data53.txt};
\end{axis}
\end{tikzpicture}  
\end{subfigure}
\caption{Evolution of key variables during the ensemble learning process for \textbf{SRCC} over the \textbf{WS353} dataset}
\label{fig:metricsff}
\end{figure}

\subsection{Summary of Results}
Table \ref{tab:sum1} summarizes the results obtained for the \textbf{MC30} benchmark dataset. Each section features two columns: the first denoting the method or ensemble used and the second representing the performance, i.e., the PCC in the initial section and SRCC in the subsequent section. These scores assess the degree of correlation between the predicted and ground truth values. Values are reported as the median of the results of the 30 independent runs. 

\begin{table}[H]
\centering
\caption{Summary of results obtained for the \textbf{MC30} benchmark dataset}
\begin{tabular}{|c|c|}
\hline
\textbf{Method/Ensemble} & \textbf{PCC}\\
\hline
Google distance \citep{key-Cilibrasi} & 0.470 \\
Huang et al. \citep{key-Huang} & 0.659 \\
J \& C \citep{key-Jiang} & 0.669 \\
Resnik \citep{key-Resnik} & \underline{0.780} \\
\hline
Bert-Cos.            & 0.740  \\
Bert-Man.            & 0.744  \\
Bert-Euc.            & \underline{0.751}  \\
Bert-Inn.            & 0.728   \\
Bert-Ang.            & 0.746   \\
\hline
LR                   & 0.757   \\
TGP                  & 0.757   \\
LGP                  & \underline{0.845}  \\
CGP                  & 0.777   \\
\textbf{GE}          & 0.794   \\
\textbf{GE-i}  & 0.752   \\
\hline
\end{tabular}
\quad
\begin{tabular}{|c|c|}
\hline
\textbf{Method/Ensemble} & \textbf{SRCC}\\
\hline
Aouicha et al. \citep{key-Aouicha}  & 0.640  \\
J \& C \citep{key-Jiang} & 0.669 \\
Lin \citep{key-Lin} & 0.619 \\
Resnik \citep{key-Resnik} & \underline{0.757} \\
\hline
Bert-Cos.            & 0.701  \\
Bert-Man.            & 0.689  \\
Bert-Euc.            & \underline{0.718}  \\
Bert-Inn.            & 0.711   \\
Bert-Ang.            & 0.701   \\
\hline
LR                   & 0.770   \\
TGP                  & 0.758   \\
LGP                  & 0.822   \\
CGP                  & 0.766   \\
\textbf{GE}          & \underline{0.859}   \\
\textbf{GE-i}  & 0.827   \\
\hline
\end{tabular}
  \label{tab:sum1}
\end{table}

The tabular presentation of the results enables comparisons of the effectiveness of various methods or ensembles, thus facilitating the identification of optimal approaches for the specific task. We can see that LGP is giving better results for \textbf{PCC} and GE for \textbf{SRCC}.

Table \ref{tab:sum2} summarizes the results obtained for the \textbf{GeReSiD50} benchmark dataset. The table also consists of two sections, each containing two columns. The first column displays the method or ensemble used in the study, while the second column represents the performance denoted as the \textbf{PCC} and \textbf{SRCC}, respectively. Values are again reported as the median result of the 30 independent runs.

\begin{table}[H]
\centering
\caption{Summary of results obtained for the \textbf{GeReSiD50} benchmark dataset}
\begin{tabular}{|c|c|}
\hline
\textbf{Method/Ensemble} & \textbf{PCC}\\
\hline
Aouicha et al. \citep{key-Aouicha}  & \underline{0.640}   \\
Deerwester et al. \citep{key-Deerwester} & 0.594 \\
Han et al. \citep{key-Han}    & 0.490 \\
Han et al. v2 \citep{key-Han}  & 0.630    \\
\hline
Bert-Cos.            & 0.725  \\
Bert-Man.            & 0.706  \\
Bert-Euc.            & 0.711  \\
Bert-Inn.            & \underline{0.735}  \\
Bert-Ang.            & 0.722  \\
\hline
LR                   & 0.736   \\
TGP                  & 0.735   \\
LGP                  & \underline{0.756}   \\
CGP                  & 0.738   \\
\textbf{GE}          & 0.743   \\
\textbf{GE-i}  & 0.735   \\
\hline
\end{tabular}
\quad
\begin{tabular}{|c|c|}
\hline
\textbf{Method/Ensemble} & \textbf{SRCC}\\
\hline
Gabrilovich \citep{key-Gabrilovich} & \underline{0.680} \\
J \& C  \citep{key-Jiang}      & 0.310  \\
Lin \citep{key-Lin} & 0.390 \\
Resnik \citep{key-Resnik} & 0.260 \\
\hline
Bert-Cos.            & 0.724  \\
Bert-Man.            & 0.715  \\
Bert-Euc.            & 0.727  \\
Bert-Inn.            & \underline{0.740}   \\
Bert-Ang.            & 0.724   \\
\hline
LR                   & 0.744   \\
TGP                  & 0.740   \\
LGP                  & 0.752   \\
CGP                  & 0.745   \\
\textbf{GE}          & \underline{0.779}   \\
\textbf{GE-i}  & 0.740   \\
\hline
\end{tabular}
  \label{tab:sum2}
\end{table}

It is possible to see that when operating over the \textbf{GeReSiD50} dataset, LGP performs better in terms of \textbf{PCC}, and GE presents better results in terms of \textbf{SRCC}, as in the previous case.

Table \ref{tab:sum3} summarizes the results obtained for the \textbf{WS353} benchmark dataset. The table is divided into two sections, each with two columns. The first column lists the methods or ensembles used in the study, while the second column shows their performance, indicated by \textbf{PCC} and \textbf{SRCC}. All values represent the median results from 30 independent runs.

\begin{table}[H]
\centering
\caption{Summary of results obtained for the \textbf{WS353} benchmark dataset}
\begin{tabular}{|c|c|}
\hline
\textbf{Method/Ensemble} & \textbf{PCC}\\
\hline
Rada et al. \citep{key-rada}  & 0.340   \\
Leacock et al. \citep{key-lch} & 0.349 \\
Wu and Palmer \citep{key-wup}    & 0.361 \\
Resnik \citep{key-Resnik}  & \underline{0.385}   \\
\hline
Bert-Cos.            & 0.810  \\
Bert-Man.            & 0.752  \\
Bert-Euc.            & 0.762  \\
Bert-Inn.            & \underline{0.811}  \\
Bert-Ang.            & 0.777  \\
\hline
LR                   & 0.262   \\
TGP                  & 0.811   \\
LGP                  & 0.817   \\
CGP                  & 0.811   \\
\textbf{GE}          & \underline{0.827}   \\
\textbf{GE-i}  & 0.811   \\
\hline
\end{tabular}
\quad
\begin{tabular}{|c|c|}
\hline
\textbf{Method/Ensemble} & \textbf{SRCC}\\
\hline
Rada et al. \citep{key-rada}  & 0.314   \\
Leacock et al. \citep{key-lch} & 0.314 \\
Wu and Palmer \citep{key-wup}    & \underline{0.348} \\
Resnik \citep{key-Resnik}  & 0.347   \\
\hline
Bert-Cos.            & 0.817  \\
Bert-Man.            & 0.792  \\
Bert-Euc.            & 0.817  \\
Bert-Inn.            & 0.817   \\
Bert-Ang.            & 0.817   \\
\hline
LR                   & 0.470   \\
TGP                  & 0.812   \\
LGP                  & \underline{0.817}   \\
CGP                  & 0.812   \\
\textbf{GE}          & \underline{0.817}   \\
\textbf{GE-i}  & 0.804   \\
\hline
\end{tabular}
 \label{tab:sum3}
\end{table}

Please note that while GE achieves higher accuracy when evolving complex ensemble structures, GE-i prioritizes interpretability by enforcing constraints on formula complexity, making it more suitable for applications requiring human-readable explanations at the cost of slight performance degradation. It is also necessary to bear in mind that while our approach demonstrates a strong correlation with human judgment, certain cases reveal its limitations. For example, the ensemble occasionally struggles with fine-grained semantic distinctions, such as differentiating between near-synonyms and context-dependent meanings. Sometimes, it also overestimates similarity for conceptually related but non-synonymous terms while underestimating strong synonymy. A deeper analysis of such cases could help refine the fitness function or introduce mechanisms for handling contextual nuances.

\subsection {Ablation Study and Sensitivity Analysis}
The parameters listed in Table~\ref{tab:param} are critical to our GE process, as variations in these settings could potentially influence performance. To assess their impact, we have conducted a sensitivity analysis focusing on key parameter choices:   

\begin{itemize}
    \item The choice of crossover method (e.g., \texttt{variable\_onepoint}) and its probability governs the exchange of genetic material between individuals. While higher probabilities can improve exploration, they may also disrupt well-performing genomes if overused. We tested probabilities between 0.6 and 0.8 but observed no significant differences in results.  
    
    \item We have increased the number of generations to 400 to provide more opportunities for evolution but have incurred higher computational costs with no benefits.  
    
    \item We have also experimented with a larger population of 200 individuals to promote genetic diversity and reduce premature convergence risks. However, this did not yield improvements in accuracy and introduced additional computational costs again.  
    
    \item Last but not least, we have adjusted mutation to introduce variability and help escape local optima. Despite our efforts, no noticeable improvements have been observed with these changes.  
\end{itemize}

We have additionally tested other parameters, including $MAX\_GENOME\_LENGTH$ and $MAX\_TREE\_DEPTH$ to control solution complexity, $INITIALISATION$ to influence genetic diversity and exclude invalid genomes, and $SELECTION$ and $REPLACEMENT$ strategies to balance exploration and exploitation. While these adjustments have not yielded measurable improvements, further exploration and fine-tuning of these parameters remain as future work.

\section{Discussion}
Semantic similarity ensembles are advantageous over other methods as they can use the capabilities of a broad spectrum of established similarity measures. As a result, these models often yield predictions of superior accuracy compared to utilizing individual methods in isolation. Our work has shown the advantages of using GE techniques to build ensembles in this context. Our research on the use of GE to address this specific challenge allows identifying several advantages over traditional GP-based strategies:

\begin{enumerate}
	\item Our approach presents greater flexibility by allowing the evolution of solutions with diverse structures, adapting dynamically to different datasets and similarity measures. Unlike traditional ensemble methods that rely on predefined aggregation rules, GE evolves customized formulas that optimize performance based on the specific characteristics of the input data.
	\item Our approach improves efficiency compared to other GP-based methods by generating directly executable code for each solution. This reduces computational overhead and speeds up the evolution process.
	\item Our approach is able to model an appropriate trade-off between accuracy and interpretability, allowing for the evolution of ensembles that are either highly accurate, interpretable, or a balanced combination of both, depending on the specific requirements of the task.
	\item Our approach can be applied to real-world tasks such as semantic search, document clustering, and chatbots, where optimizing similarity measures dynamically can improve accuracy and adaptability beyond fixed-rule approaches. Beyond semantic similarity, the proposed approach could be adapted to other NLP tasks such as text classification, paraphrase detection, and question answering, where evolving optimized aggregation strategies can improve model performance.
\end{enumerate}

However, GE also has its own drawbacks. One area for improvement lies in finding good starting points for the search process and interpreting the evolved solutions. Our approach might improve efficiency but sacrifice transparency, making it challenging for human operators to understand the evolved solutions' inner workings. While GE can optimize similarity measures, the resulting formulas can be complex and difficult to interpret. A potential solution is to introduce constraints during evolution to favor simpler expressions. Future work could explore hybrid approaches that balance accuracy with human interpretability, enabling a clearer understanding of how semantic similarity decisions are made.

Moreover, while GE performs well in optimizing semantic similarity ensembles, its computational cost is higher than traditional methods such as linear regression. The evolutionary process requires multiple iterations of fitness evaluations, making it more resource-intensive. Future work could explore techniques such as parallelized evolution or pruning strategies to improve efficiency while maintaining accuracy.

Last but not least, our approach operates with a pool of candidate similarity measures predefined and fixed throughout the evolutionary process. This restricts the adaptability of the ensemble to new tasks or domains. Future work could explore dynamic measure selection, where GE is allowed to discover and incorporate new similarity measures during evolution. This could improve the model's generalization capability and enable it to adapt to different types of textual data without requiring manual selection. Furthermore, future work should include runtime benchmarks comparing GE with fine-tuned transformer models.

\section{Conclusion}
In this work, we have presented a novel approach for the automated design of ensembles of semantic similarity measures using GE. Through empirical evaluations on several benchmark datasets, we have demonstrated the good results of our method compared to existing state-of-the-art GP-based ensembles in some cases. These findings show the potential of GE for automatic semantic similarity measure selection and aggregation when the goal is to achieve superior performance compared to individual semantic similarity measures. In addition, sustainability criteria (e.g. interpretability constraints) are taken into account that are not usually present in DL-based solutions.

Furthermore, our proposed strategy offers several notable advantages over traditional methods: It enables handling a large pool of candidate semantic similarity measures without requiring manual feature selection or parameter tuning, alleviating the process's time-consuming and knowledge-intensive aspects. Moreover, our approach demonstrates flexibility, allowing human operators to easily add or remove semantic similarity measures from the initial pool per their requirements.

Our approach shows strong potential for automating system design and could be extended to other domains beyond semantic similarity. While this strategy outperforms traditional ensemble learning techniques, its performance against fine-tuned transformer models like BERT remains unexplored. Future work should assess how GE-based ensembles compare to deep learning approaches regarding accuracy, efficiency, and interpretability. Further research could also explore the application of GE to other NLP tasks and refine search strategies and fitness functions.

\section*{Code Availability}
The data and source code that illustrates this approach is available at the following repository \url{https://github.com/jorge-martinez-gil/sesige}.

\section*{Acknowledgments}
The author thanks the anonymous reviewers for their help in improving the manuscript. The research reported in this paper has been funded by the Federal Ministry for Climate Action, Environment, Energy, Mobility, Innovation, and Technology (BMK), the Federal Ministry for Digital and Economic Affairs (BMDW), and the State of Upper Austria in the frame of SCCH, a center in the COMET - Competence Centers for Excellent Technologies Programme managed by Austrian Research Promotion Agency FFG.

%

\bibliography{mybib}

\end{document}